%% file: main.tex
\documentclass[10pt,twocolumn,letterpaper]{article}

\usepackage[pagenumbers]{cvpr} 

\usepackage{graphicx}
\usepackage{amsmath}
\usepackage{amssymb}
\usepackage{booktabs}

\usepackage[pagebackref,breaklinks,colorlinks]{hyperref}

\usepackage[capitalize]{cleveref}
\crefname{section}{Sec.}{Secs.}
\Crefname{section}{Section}{Sections}
\Crefname{table}{Table}{Tables}
\crefname{table}{Tab.}{Tabs.}

\def\cvprPaperID{*****} 
\def\confName{CVPR}
\def\confYear{2022}

\DeclareUnicodeCharacter{4E8E}{}
\DeclareUnicodeCharacter{529B}{}
\DeclareUnicodeCharacter{519B}{}

\input{preamble}
\begin{document}
\input{sec/0_metadata}

\maketitle
\input{sec/0_abstract}
\input{sec/1_introduction}
\input{sec/2_related}

\input{sec/3_method}
\input{sec/4_results}
\input{sec/5_conclusions}

{
    \small
    \bibliographystyle{ieee_fullname}
    \bibliography{main}
}



\end{document}

%% file: preamble.tex
\usepackage{overpic}
\usepackage{enumitem} 
\usepackage{overpic} 
\usepackage{color}

\definecolor{turquoise}{cmyk}{0.65,0,0.1,0.3}
\definecolor{purple}{rgb}{0.65,0,0.65}
\definecolor{dark_green}{rgb}{0, 0.5, 0}
\definecolor{orange}{rgb}{0.8, 0.6, 0.2}
\definecolor{red}{rgb}{0.8, 0.2, 0.2}
\definecolor{darkred}{rgb}{0.6, 0.1, 0.05}
\definecolor{blueish}{rgb}{0.0, 0.3, .6}
\definecolor{light_gray}{rgb}{0.7, 0.7, .7}
\definecolor{pink}{rgb}{1, 0, 1}
\definecolor{greyblue}{rgb}{0.25, 0.25, 1}

\usepackage{blindtext}

\renewcommand{\paragraph}[1]{\vspace{1em}\noindent\textbf{#1}.}

%% file: sec/0_metadata.tex
\title{ Multi-Image Visual Question Answering}

\author{
$^*$Akhilesh Bhardwaj \\ $^*$Harsh Raj \\ $^*$Janhavi Dadhania \\
Supervisor : Prof. Prabuchandran K.J.\\
Indian Institute Of Technology Dharwad\\
}

%% file: sec/0_abstract.tex
\begin{abstract}
While a lot of work has been done on developing models to tackle the problem of Visual Question Answering, the ability of these models to relate the question to the image features still remain less explored. 

\quad We present an empirical study of different feature extraction methods with different loss functions. We propose New dataset for the task of Visual Question Answering with multiple image inputs having only one ground truth, and benchmark our results on them. Our final model utilising Resnet + RCNN image features and Bert embeddings, inspired from stacked attention network gives 39\% word accuracy and 99\% image accuracy on CLEVER+TinyImagenet dataset. \\ \\
code: \href{https://github.com/harshraj22/vqa}{https://github.com/harshraj22/vqa}

\end{abstract}

%% file: sec/1_introduction.tex
\section{Introduction}
\label{sec:intro}
%
\input{fig/teaser}

Cross Modal task of visual question answering can be helpful in various areas for example visual aid to
blind people, image captcha solver. Present state of the art methods solves the task in three parts,
1) image features extraction 2) Question feature extraction and 3) merging two modalities to finally predict
the answer.
\\

\includegraphics[width=1\columnwidth]{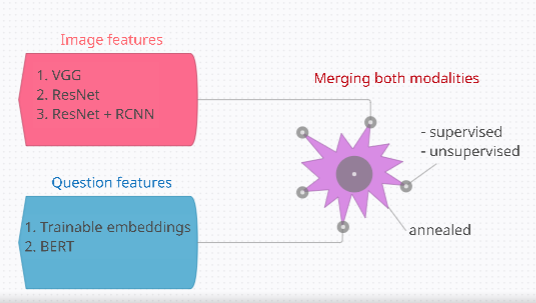}

\newpage
Good quality image and question features can improve the accuracy significantly. Few of the candidate models
for image feature extraction are VGG \cite{simonyan2014very}, ResNet\cite{he2016deep}, RCNN\cite{girshick2014rich}. LSTMs\cite{hochreiter1997long} are generally used to extract question features after
encoding the question using word embeddings. Recently BERT\cite{devlin2019bert} has proved to be an excellent text encoder which can
further improve the performance. Merging of the two modalities is usually done by applying attention. Current
state-of-the-art architectures are complex or tries to use different techniques like ensemble on top of simple
structure for example oscar, pythia.

In this work we experiment with successful image feature extractors and question encoders which in our knowledge
has not been tried yet. We also observe the significant improvement in the performance as we move from VGG to RCNNs
and trainable embeddings to BERT.

We also propose modified VQA problem statement that first tries to find out the correct image that corresponds to the
question and then answers the question.

\paragraph{Contributions}
\begin{itemize}[leftmargin=*]
\setlength\itemsep{-.3em}
\item Empirical study of different feature extraction methods and their combination for feature representation.
\item New balanced datasets.
\item Experimental results with different training techniques.
\end{itemize}

\def\thefootnote{*}\footnotetext{These authors contributed equally to this work}\def\thefootnote{\arabic{footnote}}

%% file: fig/teaser.tex


%% file: sec/2_related.tex
\input{fig/overview} 

\section{Related works}
\label{sec:related}
Show, attend, ask and answer \cite{kazemi2017show} model generalise all the previous works on VQA and claims that the simple architecture
can give state-of-the-art performance. Stacked attention network \cite{yang2016stacked} has the similar skeleton but it better utilize
attention method. RstNet \cite{zhang2021rstnet} proposes new method to ensure that spatial representation is not lost while flattening the image features. Oscar \cite{li2020oscar} uses new learning method using object tags detected in images to ease the model's learning. We use Stacked Attention Network model as our base model for below reasons,\\
1. Include general techniques used for VQA i.e CNN for image features and LSTM for question encoding and attention while merging.\\
2. Simple architecture gives satisfactory results comparable to complex ones.\\
There are many complex models (bottom up attention \cite{anderson2018bottomup}, ERNIE-VIL \cite{yu2020ernie}, Graph VQA \cite{teney2017graph}, VL-BERT \cite{su2019vl}) that show slight
improvement in performance but also has significantly more number of parameters.





%% file: fig/overview.tex
\begin{figure*}
\begin{center}
\includegraphics[width=2\columnwidth]{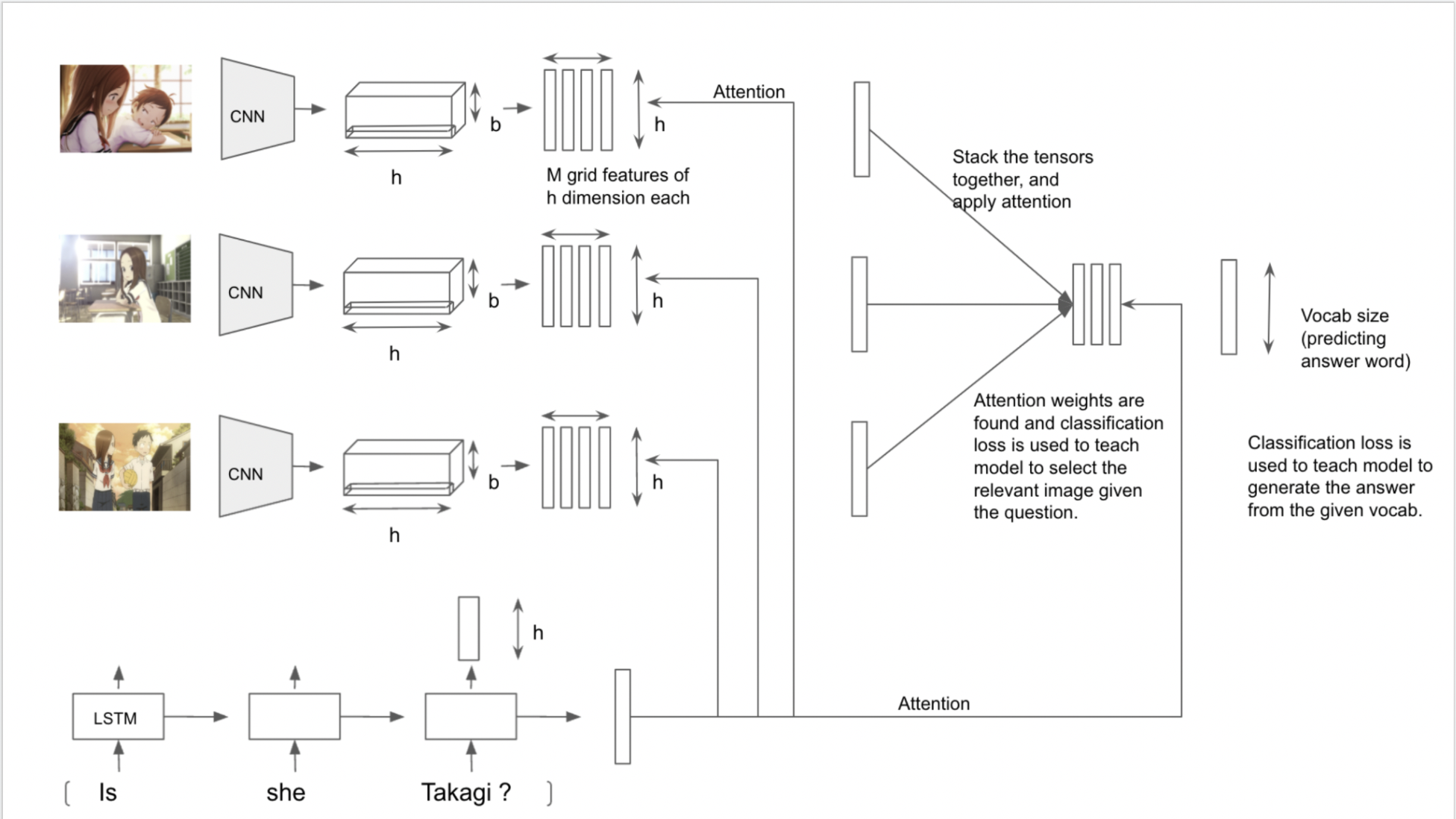}
\end{center}
\caption{
\textbf{Architecture -- }
A figure summarizing our base model
}
\label{fig:overview}
\end{figure*}

%% file: sec/3_method.tex
\section{Method}

Our proposed approach is derived from Stacked Attention Networks \cite{yang2016stacked}  which is for single image Visual Question Answering. We modified the model such a way that it can accommodate multiple images at a time and predict the answer along with the image from which it is most accurate to answer among the many input images.
Model architecture is broadly divided into three parts. 
\begin{itemize}
    \item Image Encoder: We first extract features from images using pre-trained image classification models such as VGG, ResNet and object detection models like RCNN. For the image classification models, we consider the features just before the flattening step, while for the object detection models, we use the region proposals features. The extracted features are flattened and merged to yield an output tensor of shape [-1,196,640].
    \item Question Encoder: We first tokenize the words in question to pass it through sequence model. We experiment with two different sequence models: LSTM and BERT. For LSTM, we use NLTK tokenizer while for BERT embedding we used BERT tokenizer because of their better compatibility with each other. After passing the questions through sequence model, we get question features as an output tensor of shape [-1, seq-len, 640] where is seq-len in max length of question that model can take.
    \item Attention: This part of the model works on efficiently combining the output of image and question encoder. We use multiple layers of Multi Head Attention\cite{DBLP:journals/corr/abs-2006-16362} to merge the question features with the image features. The image features are used as queries and values, while the question features are used as keys. 
\end{itemize}
Multiple input images are first passed through Image encoder while, questions are passed through Question Encoder. Lets say, for each data point in the dataset, a set of N images is input to Image encoder giving us output of [N, 196, 640] and meanwhile Question Encoder outputs [1, 30, 640] in this case seq-len is 30. Attention network takes input output of Image encoder and Question encoder and outputs weighted sum of the image features output of the Encoder. Then weighted sum image is used to answer the question in an old fashion single image visual Question Answering. Finally when Output in converted to answer vocab size using a Linear fully connected layer and converting output of Linear by passing it through softmax activation function and word with most softmax value is being output as an output of the whole model.\\
\newline
\textbf{Loss Function}: We treat Multi Image Visual Question Answering as a classification problem and use Cross Entropy loss for both Image as well as Word.
\begin{itemize}
    \item Word loss + $\lambda$ Image Loss: used a weighted average to two losses to train the network to predict the correct image to answer and then answer correctly after looking at the image.
    \item Word loss only: we may not always have liberty to have the correct image to train the model. Keeping this in mind we trained model with word loss only to try to train model to predict correct image using word loss only.
\end{itemize}

\textbf{Annealed Training}: In order to allow our model to learn the important correlations between the image features and the question features, we start the training with a high constant factor ($\lambda$) for image classification loss, and gradually decrease its value after each epoch. The model first learns to classify the correct image using the correct questions, and then to answer the question using the information from the image.

%% file: sec/4_results.tex
\input{tab/sota}
\input{tab/ablations}

\section{Experiments and Results}
\subsection{Datasets}
We propose the following two datasets in order to benchmark our model's performance.
\subsubsection{CLEVER + TinyImagenet}
CLEVER\cite{johnson2017clevr} contains 100,000 synthetically generated images containing 3 to 10 objects with random shapes, color, size, materials etc and 999,968 questions. Tiny Imagenet \cite{le2015tiny} contains 100,000 images of 200 classes (500 for each class) downsized to 64×64 colored images. 
    
\quad We use CLEVER for the ground truth image, and questions, and randomly sample 3 images from Tiny Imagenet as non-ground truth image in our dataset.

\subsubsection{VQA2.0 + Caltech256}
VQA2.0\cite{balanced_vqa_v2} is a balanced version of VQA \cite{VQA} dataset containing 265,016 images and 5.4 questions on average per image. 
We use VQA2.0 for the questions, and detect all objects present in the image using YOLOv5 \cite{glenn_jocher_2020_4154370}, and randomly sample images from Caltech256 dataset \cite{griffin2007caltech}, which do not belong to the detected classes. This ensures that the 3 non-ground truth images are equally realistic and relevant to the ground truth image.

\subsection{Results}
We benchmark our model's performance on CLEVER \cite{johnson2017clevr} + TinyImagenet \cite{le2015tiny} dataset. Our base model with pre-trained VGG19 as backbone and trainable embeddings gives 4.2\% word accuracy and 24\% image classification accuracy on being trained with a weighted sum of Image Classification loss and word classification loss.

\quad We then improve our model by adding ResNet as a better feature extractor, and Bert for pre-trained word embeddings, improving the word classification accuracy to 32\% and image classification accuracy to 99\%. Following Pythia 1.0 \cite{jiang2018pythia}, we concatenate grid features extracted by ResNet and Region Features by RCNN as input to the model. This modified model is trained with annealed method giving 39\% word accuracy and 99\% image accuracy.

%% file: tab/sota.tex
\begin{table}
\centering
\resizebox{\linewidth}{!}{ 
\begin{tabular}{@{}lccc@{}}
\toprule
Method & CLEVER + Tiny Imagenet & VQA2.0 \\ 
\midrule
VGG & 0.042 & 0.18 \\
ResNet + Bert &  0.32 & - \\
Resnet + Bert + RCNN (Annealed) & 0.39 & -\\
\bottomrule
\end{tabular}
} 
\caption{
\textbf{Word Accuracy} of different models.
} 
\label{tab:sota}
\end{table}

\begin{table}
\centering
\resizebox{\linewidth}{!}{ 
\begin{tabular}{@{}lccc@{}}
\toprule
Method & CLEVER + Tiny Imagenet \\ 
\midrule
VGG & 0.24  \\
ResNet + Bert &  0.99 \\
Resnet + Bert + RCNN (Annelead) & 0.99\\
\bottomrule
\end{tabular}
} 
\caption{
\textbf{Image Accuracy} of different models.
} 
\label{tab:sota_}
\end{table}

%% file: tab/ablations.tex
\begin{table}
\centering
\resizebox{\linewidth}{!}{ 
\begin{tabular}{@{}lccc@{}}
\toprule
Method & Without ImageLoss & With ImageLoss \\
\midrule
VGG + trainable embeddings & 0.038 & 0.042\\
ResNet + Bert + RCNN (Annealed) & 0.32 & 0.39 \\
\bottomrule
\end{tabular}
} 
\caption{
\textbf{Accuracies} with and without image loss.
} 
\label{tab:ablations}
\end{table}

%% file: sec/5_conclusions.tex
\section{Conclusions}

In this BTP we experimented with three different feature extractors i.e VGG, Resnet and RCNN and two different question encoder techniques i.e trainable embeddings and BERT, taking the stacked attention network as a base model. We experiment with different training methods and observe that,\\
1. Bert captures better features representations than training our own embeddings from scratch\\
2. RCNN features provide improvements in image's feature representation than ResNet alone\\
3. Annealed Training helps in improving accuracy even further\\
\section{Future Work}
\begin{enumerate}
    \item Benchmark all the above experiments on larger and comparatively more difficult VQA1.0 dataset that would help us understand advantages/disadvantages of multiVQA.
    \item Improve the dataset further by using existing augmentation models like Neural Style Transfer.
    \item Modify the loss function by using MSE loss for word embeddings instead of Cross-Entropy.
    \item Further modify the task to predict multiple ground truth images.
\end{enumerate}